\documentclass{article}

\usepackage{times}
\usepackage{graphicx} 
\usepackage{subfigure} 

\usepackage[numbers,compress]{natbib}

\usepackage{algorithm}
\usepackage{algorithmic}

\usepackage{hyperref}

\usepackage[accepted]{icml2017}

\usepackage{enumerate}
\usepackage[utf8]{inputenc} 
\usepackage[T1]{fontenc}    
\usepackage{hyperref}       
\usepackage{url}            
\usepackage{booktabs}       
\usepackage{amsfonts}       
\usepackage{colortbl}

\usepackage{tikz}
\usepackage{pgfplots}
\usepgfplotslibrary{groupplots} 
\usetikzlibrary{pgfplots.groupplots}
\pgfplotsset{compat = 1.3} 

\usepackage{graphicx}

\usepackage{rays_defs_13}

\newcommand\arewardT{\EX{\mathrm{reward}(T)}}

\newtheorem{lemma}{Lemma}

\newtheorem{theorem}{Theorem}
\newtheorem{example}{Example}

\newtheorem{proposition}{Proposition}

\newcommand\pf{p_{\mathrm{f}}}
\newcommand\nuser{N} 
\newcommand\nitem{M}
\newcommand\ngroups{K}

\newcommand{\vratingS}{\vocrating^{\text{sim}}}

\newcommand\ocrating{R}

\newcommand\vocrating{\vr}
\newcommand\ocR{R}
\newcommand\R{R^{\mathrm{hidden}}}

\newcommand{\setQ}{\mc Q}

\usepackage{dsfont}
\newcommand\ind[1]{\mathds{1}\left\{#1\right\}}

\renewcommand\PR[2][\Tex]{
\ifthenelse{\equal{#1}{}}{{\mathbb P}\left[#2\right]}{\ensuremath{{\mathbb P}_{#1}\left[ #2\right]}}}

\icmltitlerunning{Online One-Class Collaborative Filtering}

\begin{document}

\twocolumn[
\icmltitle{The Sample Complexity of Online One-Class Collaborative Filtering}




\begin{icmlauthorlist}
\icmlauthor{Reinhard Heckel}{to}
\icmlauthor{Kannan Ramchandran}{to}
\end{icmlauthorlist}

\icmlaffiliation{to}{University of California, Berkeley, California, USA}

\icmlcorrespondingauthor{Reinhard Heckel}{heckel@berkeley.edu}

\icmlkeywords{machine learning, collaborative filtering, one-class, sample complexity}

\vskip 0.3in
]



\printAffiliationsAndNotice{}  

\begin{abstract} 
We consider the online one-class collaborative filtering (CF) problem that consists of recommending items to users over time in an online fashion based on \emph{positive} ratings only. 
This problem arises when users respond only occasionally to a recommendation with a positive rating, and never with a negative one. 
We study the impact of the probability of a user responding to a recommendation, 
$\pf$, on the sample complexity, i.e., the number of ratings required to make `good' recommendations, and ask whether receiving positive and negative ratings, instead of positive ratings only, improves the sample complexity. 
Both questions arise in the design of recommender systems. 
We introduce a simple probabilistic user model, and analyze the performance of an online user-based CF algorithm. 
We prove that after an initial \emph{cold start phase}, 
where recommendations are invested in exploring the user's preferences, 
this algorithm makes---up to a fraction of the recommendations required for updating the user's preferences---perfect recommendations. 
The number of ratings required for the cold start phase 
is nearly proportional to $1/\pf$, 
and that for 
updating the user's preferences is essentially independent of $\pf$. As a consequence we find that, 
receiving positive and negative ratings instead of only positive ones improves the number of ratings required for initial exploration 
by a factor of $1/\pf$, which can be significant. 
\end{abstract}

\section{Introduction}

Recommender systems seek to identify the subset of a large collection of items that a user likes \cite{aggarwal_recommender_2016}. 
In practice, recommender systems often use collaborative filtering (CF) \cite{ekstrand_collaborative_2011} to identify items a given user likes, based on ratings that this user and a large number of other users have provided in the past. 
To this end, a user-based CF algorithm first identifies similar users, and then predicts the ratings of a given user from the ratings provided by similar users. 
In practice, recommender systems typically operate in an online fashion, i.e., items are recommended to users over time, and the ratings obtained in response to recommendations are used to improve future recommendations. 

However, in many application areas of recommender systems, users only occasionally rate what they `like', and never what they `dislike'. 
E.g., in e-commerce, such as Amazon's recommender system, an item (or a set of items) is recommended to a user, and the user either purchases the item, which indicates a `like', or the user does not purchase the item. 
Not purchasing the item, however, does not necessarily indicate a `dislike', since the user might not even have considered the recommendation. 
Other examples of such feedback include implicit ratings such as viewing a webpage and listening to a song~\cite{hu_collaborative_2008}, 
and business-to-business recommender systems~\cite{heckel_scalable_2016}. 
The problem of generating recommendations based on positive ratings only is known as one-class CF  \cite{pan_one-class_2008}. 
The lack of negative ratings is often considered to make this problem challenging \cite{pan_one-class_2008}.  
However, it is unclear whether it is fundamentally more difficult in the absence of negative ratings to identify the user's preferences,
in the sense that 
the sample complexity (i.e., the number of ratings required to make good recommendations) is fundamentally larger. 
Additionally, there is little theoretical understanding on how the probability of a user responding to a recommendation, $\pf$, affects the sample complexity of a one-class CF algorithm and in particular its \emph{cold start time}, i.e., the number of recommendations the algorithm needs to invest in learning the user's preferences before being able to make good recommendations. 
In this paper, we address those two questions, that turn out to be closely related. 

To this end, we introduce a probabilistic model for a one-class online recommender system and a corresponding online user-based CF algorithm, termed User-CF, and analyze its performance. 
Our model, and elements of our algorithm, are inspired by a related model and algorithm by~\citet{bresler_latent_2014} for the \emph{two-class CF} problem, i.e., for a setup where positive \emph{and} negative ratings are available. 
In a nutshell, each user in our model has a latent probability preference vector which describes the extent to which she likes or dislikes each item. Similar users have similar preference vectors. 
At a given time step $t=0,1,\ldots$, the User-CF algorithm recommends a single item to each user, typically different for each user. 
With probability specified by a corresponding preference vector, the user likes or dislikes the recommended item. 
If the user likes the item, the user rates it with probability $\pf$, and if the user does not like the item, no rating is given. 
An item that has been rated cannot be recommended again, since a rating often corresponds to consuming an item, and there is little point in, e.g., recommending a product that has been previously purchased in the past for a second time. 
While in practice the probability $\pf$ could be different for each user, for ease of presentation, we assume that $\pf$ is constant over all users. 
The goal of the User-CF algorithm is to maximize the number of recommendations that a users likes. 

The User-CF algorithm consists of an exploitation step that recommends items that similar users have rated positively, and two kinds of exploration steps; one to learn the preferences of the users and the other to explore similarity between users. 

Our main result, stated in Section \ref{sec:mainres}, guarantees that
after a certain cold start time
in which the User-CF algorithm recommends  the order of $(\log (\nuser) /\pf^2)^{\frac{1}{1-\alpha}}$ items to each user,
a fraction $1 - c/\pf$ of the remaining recommendations given by the User-CF algorithm are optimal. 
Here, $\alpha$ is a learning rate that can be chosen very close to zero, $\nuser$ is the number of users,  and $c$ a numerical constant. 
The cold start time is required to identify similar users and learn their preferences regarding a few items. 
We also show that any algorithm has to make on the order of $1/\pf^2$ recommendations before it can make good recommendations, therefore the User-CF algorithm is near optimal. 
The fraction $c/\pf$ of the remaining time steps is associated with learning the preferences of the users. 
This `cost' of $c/\pf$ does not have to be paid upfront, but is paid continuously: After the cold start time, the User-CF algorithm starts exploiting successfully. 
Again, a fraction of the recommendations proportional to $1/\pf$ is necessary to learn the preference of the users. 
Our numerical results in Section \ref{sec:numres} show that even if our data is not generated from the probabilistic model, but is based on real data, the cold start time and the fraction of time steps required to learn the preferences of the users are nearly proportional to $1/\pf^2$ and $1/\pf$, respectively. 

As a consequence of 
this result, we find that obtaining positive and negative ratings instead of only positive ones, improves the number of ratings required for the initial cold start period by a factor of $\pf$. 
To see this, note that the expected number of ratings obtained by a user 
in a given number of time steps or equivalently after a given number of recommendations is proportional to $\pf$. 
Thus, the number of ratings required for the initial cold start time is inversely proportional to $\pf$, and the number of ratings required for continuously learning the preferences is independent of $\pf$. 
Since $\pf=1$ corresponds to users giving positive and negative feedback (no positive feedback implies dislike when $\pf=1$), 
the number of ratings required for the cold-start time is by a factor of $1/\pf$ larger than the number of ratings required by a user-based CF algorithm that obtains positive and negative ratings. 

Those findings are relevant for the design of recommender systems, 
since both $\pf$ and whether positive, or negative and positive ratings are obtained can often be incorporated in the design of a recommender system. 
Therefore an understanding of the associated benefits and costs in terms of sample complexity, as provided in this paper, is important. 
We finally note that the goal of this paper is not to improve upon state-of-the art algorithms, but rather to 
inform the design of algorithms and what to expect in terms of sample complexity as a function of the various parameters involved.



\vspace{-0.2cm}
\paragraph{Related literature:}
While to the best of our knowledge, this is the first work that \emph{analytically} studies one-class CF in an online setting,
theoretical results have been established for the two or multiple class CF problems. 
One of the first analytical results on user-based CF algorithms, an asymptotic performance guarantee under a probabilistic model, was established by \citet{biau_statistical_2010}. 
Most related to our approach is the Collaborative-Greedy algorithm studied by \citet{bresler_latent_2014} for the online two-class CF problem. 
The Collaborative-Greedy algorithm differs from our User-CF algorithm in selecting the nearest neighbors based on thresholding similarity, instead of selecting the $k$ most similar users, and in the way preferences of the users are explored. 
This difference in the exploration steps is crucial for establishing that after the cold start period, our User-CF algorithm makes optimal recommendations in a fraction $1 - c/\pf$ of the remaining time steps. 
\citet{dabeer_adaptive_2013} studies a probabilistic model in an online setup, and 
\citet{barman_analysis_2012} study a probabilistic model in an offline setup, and state performance guarantees for a two-class user-based CF algorithm. 
%
Closely related to user-based CF is item-based CF.  
Item-based CF exploits similarity in item space by 
recommending items similar to those a given user has rated positively in the past. 
Our results do not extend trivially to item-based CF, since a corresponding analysis requires assumptions on the similarity in item space, and additionally the exploration strategies of item-based CF algorithms are considerably different. 
We do not discuss item based CF algorithms 
here, 
but refer to~\cite{bresler_regret_2015} for a recent analysis of an item based CF algorithm for the two-class CF problem. 
%
Next, we note that \citet{deshpande_linear_2012} study recommender systems in the context of multi-armed bandits \cite{bubeck_regret_2012}. 
Specifically, \citet{deshpande_linear_2012} consider a model where the (continuous) ratings are described by the inner product of a user and item feature vector, and assume the item feature vectors to be given. 

A conceptually related online learning problem are multi-armed bandits with dependent arms \cite{pandey_multi-armed_2007}. 
Specifically, in this variant of the multi-armed bandit problem, the arms are grouped into clusters, and the arms within each cluster are dependent. 
The assignments of arms to clusters are assumed known. 
In our paper, we assume that users cluster in user types that have similar distributions. 
Therefore, the learning problem in our paper can be viewed as an multi-armed bandit problem with dependent arms, but the assignment of the arms to clusters is unknown. 

Finally, we note that a class of learning problems reminiscent to that considered here is partial monitoring~\cite{bartok_partial_2014}. 
While partial monitoring has been studied in the context of recommender systems~\cite{kveton_cascading_2015}, we are not aware of papers on partial monitoring in collaborative filtering. 

\vspace{-0.15cm}

\paragraph{Outline:}
In Section \ref{sec:model}, we formally specify our model,  motivate it, and state the CF problem. 
Sections \ref{sec:algformal} and \ref{sec:mainres} contain the User-CF algorithm and corresponding performance guarantee, respectively. 
In Section \ref{sec:numres} we provide numerical results on real data.
The proof of our main result can be found in the supplementary material. 

\vspace{-0.2cm}
\section{\label{sec:model}Model and learning problem}
\vspace{-0.2cm}

In this section we introduce the probabilistic model and learning problem considered in this paper. 
As mentioned previously, this model is inspired by that in \cite{bresler_latent_2014} for the two-class CF problem. 
%
\paragraph{Model:}
Consider $\nuser$ users and $\nitem$ items. 
A user may like an item ($+1$), or dislike an item ($-1$). 
Associated with each user is an (unknown) latent preference vector $\vp_u \in [0,1]^M$ whose entries $p_{ui}$ are the probabilities of user $u$ liking item $i$. 
We assume that an item $i$ is either ``likable'' for user $u$, i.e., $p_{ui} > 1/2 + \Delta$, for some $\Delta \in (0,1/2]$, or ``not likable'', i.e., $p_{ui} < 1/2 - \Delta$. 
The hidden ranking $\R_{ui}$ 
is obtained at random as $\R_{ui} = 1$ (like) with probability $p_{ui}$, and $\R_{ui} = -1$ (dislike) with probability $1 - p_{ui}$. 
The ratings are stochastic to model that users are not fully consistent 
in their rating; the parameter $\Delta$ quantifies the inconsistency (or uncertainty or noise). 
The one-class aspect is incorporated in our model by assuming that users \emph{never} reveal that they \emph{dislike} an item. 

Specifically, an CF algorithm operates on the model as follows. At each time step $t=0,1,\ldots$ the algorithm recommends a single item $i = i(t,u)$ to each user $u$---typically this item is different for each user---and obtains an realization of the binary random variable 
\[
\ocR_{ui} =  
\begin{cases}
Z_{ui} \sim \text{Bernoulli($\pf$)}, & \text{if } \R_{ui}=1,\\
0, & \text{if } \R_{ui} = -1
\end{cases}
\]
in response, independently across $u$ and $i$. 
It follows that 
$
\PR{\ocR_{ui} = 1} = p_{ui} \pf
$
and
$
\PR{\ocR_{ui} = 0} = 1 - p_{ui} \pf,
$. 
Here, $\pf$ corresponds to the probability of a user reporting a positive rating. 
As mentioned before, while one might treat the slightly more general case of the probability $\pf$ being different for each user, for ease of presentation, we assume that it is constant over the users. 
If $\ocR_{ui} = 1$, user $u$ consumes item $i$, and $i$ will not be recommended to $u$ in subsequent time steps. 
Note that $\ocR_{ui}=0$ means that either user $u$ does not like item $i$ ($\R_{ui}=-1$), or user $u$ did not respond to the recommendation. 
Therefore, if $\ocR_{ui}=0$, $i$ may be recommended to $u$ again in subsequent time steps. 
Finally, observe that if $\pf=1$, the user provides positive \emph{and} negative ratings, since $\R_{ui}=0$ implies $\ocR_{ui}=-1$ if $\pf=1$. 
%
%
%
%

\begin{figure}
\begin{center}
\includegraphics[width=4.3cm]{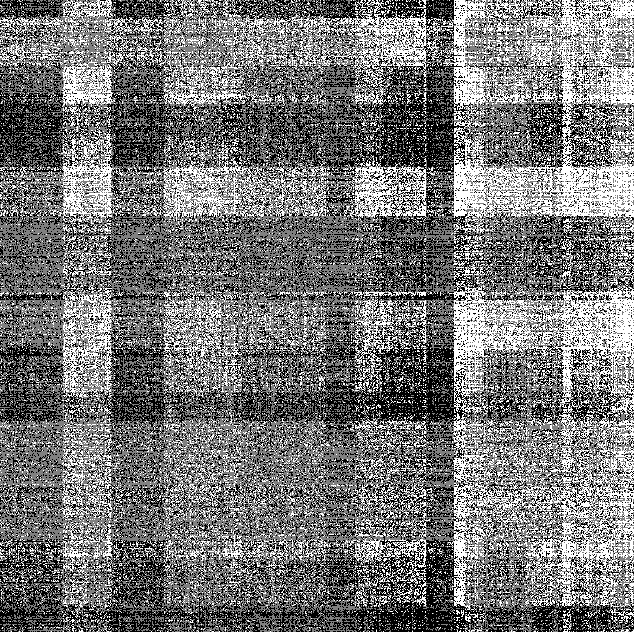}
\end{center}
\caption{
\label{fig:clusteringmovielens}
The user-items rating matrix consisting of a subset of the Movielens 10 Million dataset corresponding to the 1000 most rated movies (columns) and the 1000 users (rows) that rated most movies. 
The Movielens dataset consists of 10 Million movie ratings in $\{1,2,3,4,5\}$; 
we took the ratings $\geq 4$ as $1$, ratings $\leq 3$ as $-1$, and missing ratings as $0$; 
depicted in black, white, and gray.
}
\end{figure}

In order to make recommendations based on the user's preferences, we must assume some relation between the users. 
Following~\cite{bresler_latent_2014}, we assume that each user belongs to one of $K < N$ user types. 
Two users $u$ and $v$ belong to the same type if they find the same items likable, i.e., if 
$
\ind{ p_{ui} > 1/2 + \Delta } 
= 
\ind{ p_{vi} > 1/2 + \Delta }
$, for all items $i$. 
This does not require the preference vectors $\vp_u$ and $\vp_v$ of two users corresponding to the same user type to be equivalent. 
We note that this assumption could be relaxed by only assuming that users of the same type share a large fraction of the items that they find likable. 
We assume that 
the preference vectors belonging to the same type are more similar than those belonging to other types. 
Specifically, assume that for all $u \in [N], [N]\defeq \{0,1,\ldots,N-1\}$, and for some $\gamma \in [0,1)$, 
\begin{align}
\gamma \min_{ v \in \mc T_u }  \innerprod{\vp_u }{\vp_v}
\geq 
 \max_{v \notin \mc T_u }\innerprod{\vp_u }{\vp_v}, 
\label{eq:userssuffdiffthm}
\end{align}
where $\mc T_u \subset [N]$ is the subset of all users that are of the same type as $u$. 
The smaller $\gamma$, the more distinct users of the same type are from users of another type. 
We further assume that each user 
likes at least a fraction $\nu$ of the items. 
This assumption is made to avoid degenerate situations were a user $u$ does not like any item. 
Assuming that users cluster in the user-item space in different user types is common and is implicitly used by user-based CF  algorithms \cite{sarwar_analysis_2000}, which perform well in practice. 
To further justify this assumption empirically, we plot in Figure \ref{fig:clusteringmovielens} the clustering of user ratings of the Movielens 10 Million dataset \cite{harper_movielens_2015}. 
Figure \ref{fig:clusteringmovielens} shows that the user's ratings cluster both in user and in item space. 

\paragraph{Learning problem and reward:}
The goal of a CF algorithm is to maximize reward. 
A reasonable reward for the online CF problem is the expected number of recommendations that a user rates positively, i.e., the pseudo-reward
\[
\sum_{t=0}^{T-1} \sum_{u=0}^{N-1} \EX{R_{ui(u,t)}}.
\]
Here, $i(u,t)$ is the item recommended to $u$ at time $t$. 
In an e-commerce setting this corresponds to the number of recommended products that a user buys.
Note that due to the uncertainty of a user liking an item (the random rating $\R_{ui}$ might be $-1$ even when $i$ is likable by $u$), we cannot expect to do better than maximizing the pseudo-reward. 

In this paper our focus is on recommending likable items. Following~\cite{bresler_latent_2014} we therefore consider the closely related accumulated reward defined as the expected total number of likable items ($p_{ui}>1/2$) that are recommended by an algorithm up to time $T$:
\begin{align}
\arewardT \defeq \sum_{t=0}^{T-1} \sum_{u=0}^{N-1} \EX{X_{u i(u,t)}}. 
\label{eq:defareward}
\end{align}
Here, $X_{u i(u,t)} = \ind{p_{ui} > 1/2}$ is the indicator random variable that is equal to one if item $i(u,t)$ recommended to user $u$ at time $t$ is likable and zero otherwise (note that item $i$ is chosen by the CF algorithm as a function of the responses $\ocR_{u'i'}$ to recommendations $(i',u')$ 
made at previous time steps, and is therefore a random variable).

\section{
\label{sec:algformal}
User-CF algorithm
}
In this section we present our user-based CF algorithm (User-CF). 
In order to maximize reward the User-CF algorithm balances exploring, i.e., learning about the users, and exploiting, i.e., recommending items predicted to be likable based on previous ratings. 
To this end, the User-CF algorithm, formally introduced below, 
performs at time $t=0,1,\ldots$, either an \emph{preference exploration}, \emph{similarity exploration}, or \emph{exploitation step}. 

An \emph{exploitation step} first identifies the $k$ most similar users 
in terms of their \emph{rating vectors} $\vocrating_v \in \{0,1\}^\nitem$, for a given user $u$.  
The rating vectors consist of the responses $\ocR_{ui}$ of users to recommendations $(u,i)$ made by the User-CF algorithm at previous time steps. 
The exploitation step proceeds by recommending the item that has received the largest number of positive ratings from the nearest neighbors of $u$ in previous time steps. 
For an exploitation step to be successful, it is crucial to find similar users and learn their preferences effectively.  
This is accomplished with \emph{similarity exploration} steps, that recommend the same items to all users, and \emph{preference exploration} steps that recommend random items to certain subsets of the users. 
Before formally stating the User-CF algorithm, we illustrate its main steps using a toy example. 
 
\begin{example}
\newcommand\cg{\cellcolor{gray!15}}
Consider $N=6$ users and $M=5$ items, with preference vectors $\vp_u$
and rating vectors $\vocrating_u$ at time $t=2$ given by 
\begin{align*}
\begin{bmatrix}
\transp{\vp}_0 \\
\transp{\vp}_1 \\
\transp{\vp}_2 \\
\transp{\vp}_3 \\
\transp{\vp}_4 \\
\transp{\vp}_5 
\end{bmatrix}
=
\left[
\begin{array}{ccccc}
\cg .9 & \cg .8 & \cg .9 & .1 & .1 \\
\cg .9 & \cg .8 & \cg .9 & .2 & .3 \\
\cg .9 & \cg .8 & \cg .9 & .1 & .2 \\
.1 & .3 & .1 & \cg .8 & \cg .7 \\
.2 & .2 & .1 & \cg .9 & \cg .9 \\
.1 & .1 & .3 & \cg .7 & \cg.8 \\
\end{array}
\right], \\
\begin{bmatrix}
\transp{\vocrating}_0 \\
\transp{\vocrating}_1 \\
\transp{\vocrating}_2 \\
\transp{\vocrating}_3 \\
\transp{\vocrating}_4 \\
\transp{\vocrating}_5 
\end{bmatrix}
=
\begin{bmatrix}
\tikz \node (r1) at (0,-0.5cm) [shape=rectangle,draw,inner sep =0.05cm,outer sep=-0.1cm] {$1$}; & 0 & 0 & 0 &  \tikz \node (r1) at (0,-0.5cm) [shape=circle,draw,inner sep =0.03cm] {$0$}; \\
\tikz \node (r1) at (0,-0.5cm) [shape=rectangle,draw,inner sep =0.05cm] {$1$}; & \tikz \node (r1) at (0,-0.5cm) [shape=circle,draw,inner sep =0.03cm] {$1$}; & 0 & 0 & 0 \\
\tikz \node (r1) at (0,-0.5cm) [shape=rectangle,draw,inner sep =0.05cm] {$1$}; & \tikz \node (r1) at (0,-0.5cm) [shape=circle,draw,inner sep =0.03cm] {$1$}; & 0 & 0 & 0 \\
\tikz \node (r1) at (0,-0.5cm) [shape=rectangle,draw,inner sep =0.05cm] {$0$}; & 0 & \tikz \node (r1) at (0,-0.5cm) [shape=circle,draw,inner sep =0.03cm] {$1$}; & 0 & 0 \\
\tikz \node (r1) at (0,-0.5cm) [shape=rectangle,draw,inner sep =0.05cm] {$0$}; & 0 & 0 & \tikz \node (r1) at (0,-0.5cm) [shape=circle,draw,inner sep =0.03cm] {$1$}; & 0 \\
\tikz \node (r1) at (0,-0.5cm) [shape=rectangle,draw,inner sep =0.05cm] {$0$}; & 0 & 0 & \tikz \node (r1) at (0,-0.5cm) [shape=circle,draw,inner sep =0.03cm] {$0$}; & 0 
\end{bmatrix}.
\end{align*}
Users $0,1,2$ are of the same type as they find items $0,1,2$ likable, and users $3,4,5$ belong to a second type as they find items $3,4$ likable. 
The preference vectors are obtained by executing at time step $0$ a preference exploration step, that recommends the randomly chosen items $4,1,1,2,3,3$ to users $0,1,\ldots,5$, respectively, 
and a similarity exploration step that recommends item $0$ to all users. 
The responses $\ocR_{ui}$ obtained from the similarity and preference exploration step are marked with rectangles and circles, respectively. 
Consider recommending an item to user $0$ with an exploitation step at time $2$. The $k=2$ nearest neighbors of $u=0$ are $\mc N_u = \{1,2\}$ ($\innerprod{\vocrating_0}{\vocrating_v} = 1$ for $v=1,2$ and $\innerprod{\vocrating_0}{\vocrating_v} = 0$ for $v=3,4,5$). Since $\sum_{v \in \mc N_u} \vocrating_{vi}$ is maximized for $i=1$ ($\vocrating_{vi}$ is the $i$-th entry of $\vr_u$), item $1$ is recommended to user $u$, which happens to be a likable item, as desired. 
\end{example}

We next formally describe the User-CF algorithm, and explain the intuition behind the specific steps. 
Input parameters of the User-CF algorithm are learning rates $\alpha$ and $\eta \in (0,1)$ (e.g., $\eta=1/2$)  relevant for similarity and preference exploration steps, respectively, a batch size $Q$ relevant for preference exploration steps, and finally the number of nearest neighbors $k$, relevant for exploitation steps. 
Our results guarantee that for a range of input parameters that depends on properties of the model such as the number of user types and $\pf$, the User-CF algorithm performs essentially optimally. 
While we may or may not have prior knowledge of those model parameters, in practice, we can optimize for the hyper-parameters of the User-CF algorithm by using cross validation. 

At initialization, the User-CF algorithm generates a random permutation $\pi$ of the items $[M]$ required by the similarity exploration step. 
Furthermore, it splits the item space into $M/Q$ random, and equally sized\footnote{We assume for simplicity that $\nitem$ is divisible by $Q$, if this is not the case, batch $\lceil M/Q \rceil-1$ may simply contain less than $Q$ items.}
subsets of cardinality $Q$ (batches), denoted by $\setQ_q \subset [M]$, $q=0,\ldots,M/Q-1$. 

At time steps $t = \lfloor \eta Q q \rfloor, q = 0,\ldots,M/Q-1$, the User-CF algorithm performs preference exploration steps. 
At all other time steps, with probabilities 
$
p_J = \frac{1}{(t-q)^\alpha}, q = \lfloor t / (\eta Q)\rfloor$, and $p_E = 1- p_J
$, the algorithm performs similarity exploration and exploitation steps, respectively. 

{\bf Similarity exploration step}: For each user $u$, recommend the first item $i$ in the permutation $\pi$ that has not been recommended to user $u$ in previous steps of the algorithm. 
This step explores the item space and its important for selecting `good' neighborhoods. Performing a sufficient number of similarity exploration steps allows to guarantee that the nearest neighbors of a given user $u$ are of the same type. 
%

{\bf Preference exploration step:}
At time $t = \lfloor \eta Q q \rfloor$, recommend to each user $u$ an item, chosen independently and uniformly at random from $\setQ_q$, that has not been rated by $u$ in previous time steps. 
This step is important to learn the preferences of users. 
%

{\bf Exploitation step:} 
For all users $u$, estimate the probability of $u$ liking a given item $i$ as 
\begin{align}
\hat p_{ui} = 
\begin{cases}
\frac{1}{n_{ui}}   \sum_{v \in \mc N_u} \ocrating_{vi}
, &  \text{if } n_{ui}>0 ,   \\
0, 
& \text{if } n_{ui} = 0.
\end{cases}
\label{eq:defhatpui}
\end{align}
Here, $\ocrating_{vi}$ is the rating of user $v$ for item $i$ obtained in previous time steps (we use the convention $\ocrating_{vi} = 0$ if no rating was obtained at a previous time step).
Next, $\mc N_u$ is the set of users corresponding to the $k$ largest values of 
$
\innerprod{\vratingS_u}{\vratingS_v}, v \in [N]
$. 
Here, $\vratingS_{u} \in \{0,1\}^M$ is the vector containing only the responses $\ocR_{ui}$ of user $u$ to recommendations $i$ given in previous \emph{similarity} exploration steps up to time $t$, and is zero otherwise. 
Moreover, $n_{ui}$ is the number of users in $\mc N_u$ that received recommendation $i$. 
Finally, 
for each user $u$, recommend an item $i$ that maximizes $\hat p_{u i'}$ over all items $i'$ that have not been rated yet. 

The idea behind the User-CF algorithm is as follows. 
An exploitation step recommends likable items to $u$ if  $a)$ most of the neighbors of $u$ are of the same user type as $u$, and if $b)$ the items are sufficiently well explored so that $\hat p_{ui}$ indicates whether $i$ is likable by $u$ (i.e., $p_{ui}>1/2$) or not, for all $i$. 
A large portion of the first few steps is likely to be spent on similarity exploration. This is sensible, as we need to ensure that $a)$ is satisfied in order to make good recommendations. 
As time evolves, the User-CF algorithm randomly explores batches  of items, one batch at a time, in order to estimate the preferences of the users regarding the items in the corresponding batches. 
Note that if the User-CF algorithm would explore the entire item space at once, e.g., by recommending an item chosen at random from all items, 
the time required for the algorithm to make `good' recommendations would grow linearly in $\nitem$, and $\nitem$ might be very large. 
By splitting the item space into batches $\setQ_q$, the User-CF algorithm can start exploiting without having learned the preferences of the users regarding \emph{all} items. 
Finally note that the User-CF algorithm may recommend the same item at several time steps (unless the item is rated by the user); this is sensible in particular if $\pf$ is small. 

%
%

\vspace{-0.2cm}
\section{\label{sec:mainres}Main result}
\vspace{-0.2cm}

Our main result, stated below, 
shows that 
after a certain cold start time, the User-CF algorithm produces essentially optimal recommendations. 

\begin{theorem}
\label{thm:mainspecial}
Suppose that there are at least $\frac{N}{2K}$ users of the same type, for all user types, 
and that condition~\eqref{eq:userssuffdiffthm} holds for some $\gamma \in [0,1]$, which ensures that user types are distinct. 
Moreover, assume that at least a fraction $\nu$ of all items is likable to a given user, for all users. 
Pick $\delta >0$ and suppose that there are sufficiently many users per user type:
\begin{align}
\label{eq:kappacond}
\frac{N}{K} \geq \frac{c}{\nu \pf \Delta^2} \log(M/\delta) \log(4/\delta).
\end{align}
Set \begin{align*}
&T_{\text{start}} \defeq \\
&\left(
\frac{\tilde c \log(N/\delta) 
}{
\pf^2  (1 - \gamma)^2 \nu) 
} 
\right)^{\frac{1}{1-\alpha}} 
\left(1 - \max\left( \frac{1}{T} , \frac{K}{N} \frac{c \log(M/\delta)}{ \pf \Delta^2} \right) \right),
\end{align*}
where $\tilde c$ is a numerical constant. 
Then, for appropriate choices\footnote{Specifically, $\eta = c_1 \nu, 
\quad
k = c_2 \frac{N}{K},
\quad 
\text{and}
\quad 
Q = c_3  
\frac{k \pf \Delta^2 }{\log( M / \delta)}$, for numerical constants $c_1,c_2$, and $c_3$.
}
of the parameters $\eta,k$, and $Q$, 
the expected reward accumulated by the 
User-CF algorithm up to time $T \in [T_{\text{start}} , \frac{4}{5} \nu \nitem \pf]$ satisfies
\begin{align}
\frac{\arewardT}{NT}  &\geq \left( 1 - \frac{T_{\text{start}} + 1}{T} 
-  2^\alpha \frac{ (T - T_{\text{start}})^{1-\alpha} }{T(1-\alpha)} \right. \nonumber \\
&\hspace{1.1cm}\left.- \frac{K}{N} \frac{c \log(M/\delta)}{ \pf \Delta^2} \right) (1-\delta). 
\label{eq:lbonregretorig}
\end{align}
\end{theorem}



Theorem \ref{thm:mainspecial} states that after an initial cold start time on the order of $T_{start}$, 
the User-CF algorithm recommends only likable items up to a fraction $\frac{K}{N} \frac{c \log(M/\delta)}{ \pf \Delta^2}$ of the time steps. 
This follows since a oracle that only recommends likable items obtains an reward of $\arewardT = NT$. 
This yields the claim from the introduction, that after the cold start time, a fraction $1 - c/\pf$ of all recommendations made by the User-CF algorithm are likable. 
Note that condition~\eqref{eq:kappacond} allows the number of user types, $K$,
to be near linear in the number of users, $N$. 

We note that the particular choice of the parameters of the User-CF algorithm in Theorem~\ref{thm:mainspecial} is mainly out of expositional convenience; 
the supplementary material contains a more general statement. 

%
Theorem~\ref{thm:mainspecial} is proven by showing that after the initial cold start time, exploration steps recommend likable items with very high probability.  
An exploration step recommends a likable item to user $u$ provided that 

\vspace{-0.25cm}
\begin{enumerate}[a)]
\item most of the nearest neighbors of $u$ are of the same user type, and 
\item the items are sufficiently well explored so that the maximum of $\hat p_{ui}$ over items not rated yet  corresponds to a likable item. 
\end{enumerate}
\vspace{-0.25cm}

For a), we use that the user types are sufficiently distinct and that most of the nearest neighbors of $u$ are of the same user type. 
The former is ensured by condition~\eqref{eq:userssuffdiffthm}, and the latter holds after the initial cold start time which ensures that sufficiently many similarity exploration steps have been executed. 
After the initial cold start time, most of the nearest neighbors of $u$ are of the same user type, and 
essentially no further cost is required to learn the  neighborhoods. 
This is reflected in the lower bound~\eqref{eq:lbonregretorig}, by the terms depending on $T$ becoming negligible as $T$ becomes large compared to $T_{\text{start}}$. 
Note the dependence of $T_{\text{start}}$ on $\gamma$; the more similar the user types are (i.e., the closer $\gamma$ is to $1$), the longer it takes till User-CF is guaranteed to find good neighborhoods. 

\vspace{-0.2cm}
\subsection{Dependence on $\pf$ is nearly optimal}
\vspace{-0.2cm}

The cold start time of the User-CF algorithm guaranteed by Theorem~\ref{thm:mainspecial} is proportional to $(1/\pf^2)^{\frac{1}{1-\alpha}}$. 
For small learning rates $\alpha$, this scaling can not be improved significantly, as the following results shows.

\begin{proposition}
\label{prop:necessity}
Suppose that there are more items than user types, i.e., $M \geq K$. 
Fix $\lambda \in (0,1)$. 
Then there is a set of users with at least $\frac{N}{2K}$ users of the same type, for each user type, 
with preference vectors such that for all 
$T \leq \frac{\lambda}{\pf^2}$, 
the expected reward of any online algorithm is upper bounded by 
$\frac{\arewardT}{TN} \leq \lambda + \frac{1}{K}$. 
\end{proposition}
 
Proposition~\ref{prop:necessity} shows that if the cold start time is significantly smaller than $1/\pf^2$, then there are problem instances for which any algorithm mostly recommends non-likable items. 
The proposition is a consequence of the fact that after making on the order of $1/\pf^2$ recommendations, for many users we did not obtain any rating. 
A consequence of the proposition is that the cold start time of the User-CF algorithm is near optimal. 


Recall that even after the initial cold start time, a constant fraction of 
$\frac{K}{N} \frac{c \log(M/\delta)}{ \pf \Delta^2}$ of the recommendations might be non-likable. 
This fraction is the cost for establishing b). 
Specifically, in order to ensure b), the 
User-CF algorithm needs to recommend sufficiently many items to the $k$ neighbors of $u$ so that $\hat p_{ui}$ indicates whether user $u$ likes item $i$ or not, or more precisely such that the maximum of $\hat p_{ui}$ over items not rated yet corresponds to a likable item. 
This is established by showing that $\hat p_{ui} > \pf /2$ for all likable items $i \in \setQ_q, q=0,\ldots, \frac{t}{\eta Q}-1$, and $\hat p_{ui} < \pf /2$ for all other items. 
Since the expected number of positive ratings 
per recommendation is proportional to $\pf$, the number of ratings required to ensure that $p_{ui}>\pf/2$ is proportional to $1/\pf$. 





\vspace{-0.2cm}
\subsection{One versus two class CF}
\vspace{-0.2cm}

Recall that $\pf=1$ implies that users provide positive and negative ratings, and that the User-CF algorithm is nearly optimal in $\pf$. Since the expected number of ratings obtained by a user in a given number of time steps is proportional to $\pf$, 
 a consequence of our result is that receiving positive and negative ratings instead of only positive ones improves the number of ratings required for initial exploration by a factor of $1/\pf$, which can be significant. 

We finally note that \citet{bresler_latent_2014} proved a performance guarantee for a closely related two-class collaborative CF algorithm termed Collaborative-Greedy. \citet{bresler_latent_2014} consider the regime where the number of users is much larger than the number of items, i.e., $\nuser = O(\nitem^C), C > 1$ and additionally the number of user types obeys ($\nuser = O(\ngroups \nitem)$). 
For this regime, Theorem~\ref{thm:mainspecial} particularized to $\pf=1$ essentially  reduces to Theorem 1 in \cite{bresler_latent_2014} (there are some further minor differences). 
However, our result particularized to the two-class case also holds when the number of items is much larger than the number of users, and allows the number of user types to be near linear in the number of users. 
This improvement is due to differences in the preference exploration strategies of the algorithms.

\vspace{-0.2cm}
\subsection{Alternative exploration strategies}
\vspace{-0.2cm}

While there are other sensible preference exploration strategies, the essential element of our approach is to split up the item space into subsets of items $\setQ_0,\setQ_1,\ldots$, start by exploring $\setQ_0$, then allow for exploitation steps, continue with exploring $\setQ_1$, again allow for exploitation steps and so forth. 
If one explores instead the whole item space $[M]$ at the beginning, the learning time required for $\hat p_{ui}$ to indicate whether an item is likable or not, is proportional to $\nitem$, and can therefore be very large. 
To see this, consider a preference exploration step that recommends a single item to all users, chosen uniformly at random from the set of all items $[M]$. 
The expected number of  ratings obtained by executing $T_r$ such preference exploration steps  
relevant for estimating whether $p_{ui}>1/2$, is the expected number of neighbors of $u$ to which $i$ has been recommend, and is therefore proportional to $T_r k / M$. To ensure that this expectation is larger than $0$, $T_r$ has to be on the order or $\nitem$ (provided that the other parameters are fixed).

\vspace{-0.3cm}
\section{\label{sec:numres}Numerical results}
\vspace{-0.3cm}
\newcommand\RML{\mR_{\textrm{ML}}}

In this section, we simulate an online recommender system based on real-world data in order to understand whether the User-CF algorithm behaves as predicted by Theorem \ref{thm:mainspecial}, even when the data is not generated by the probabilistic model, but is based on real data. 
While an ideal dataset to validate our algorithm would consist of the ratings from all users for all items, the vast majority of ratings in standard CF datasets such as the Netflix or Movielens dataset (consisting of movie ratings) are unknown. 
To obtain a dataset with a higher proportion of ratings, following \cite{bresler_latent_2014}, we consider the subset of the Movielens dataset corresponding to frequently rated items and to users that have rated many items. 
The Movielens dataset consists of 10 Million movie ratings in $\{1,2,3,4,5\}$; we took the ratings $\geq 4$ as $1$, ratings $\leq 3$ as $-1$, and missing ratings as $0$. 
Many of the items (movies) in the dataset have a significant bias towards a positive or negative rating. 
To make sure our results are not due to exploiting such biases, we only select frequently rated items out of the (approximately) unbiased items. 
Note that a nearest neighbor based algorithm, like the User-CF algorithm, performs well in case the item ratings are very biased even when the neighborhoods are randomly selected. 
Of the resulting dataset, denoted by $\RML \in \{-1,0,1\}^{1000\times 500}$, $18.1\%$ of the ratings are $1$, $17.1\%$ are $-1$, and the remaining ones are unknown and therefore set to $0$. 

\paragraph{One class versus two class CF:}
We start with comparing the qualitative behavior of the User-CF algorithm to a two-class version of the User-CF algorithm. 
The two-class version of the User-CF algorithm differs from the one-class version in taking into account the negative ratings. Specifically, the ratings $\ocrating_{vi}$ in \eqref{eq:defhatpui} for the two-class version or the User-CF algorithm are set to $-1,0$, and $1$, if a negative, none, or a positive rating was obtained as a response to a recommendation. 
We performed the following experiment. 
If the User-CF algorithm recommends item $i$ to user $u$, it obtains the rating $\ocrating_{ui} = 1$ in response provided that $[\RML]_{ui}=1$ ($[\RML]_{ui}$ denotes the $(u,i)$-th entry of $\RML$), and $\ocrating_{ui} = 0$ otherwise, while the two-class User-CF algorithm obtains $\ocrating_{ui}=[\RML]_{ui}$ in response. 
We allow both algorithms to only recommend an item to a given user once, so after $M=500$ time steps, all items have been recommend to all users. 
We measure performance in terms of the accumulated reward, defined as
\begin{align*}
\text{acc-reward}(T) \defeq \sum_{t=0}^{T-1} \mathrm{reward}(t), 
\end{align*}
\begin{align}
\label{eq:defrewardnr} 
\mathrm{reward}(t) \defeq \frac{1}{N} \sum_{u=0}^{N-1} [\RML]_{u i(u,t)},
\end{align}
where $i(u,t)$ is the item recommended to user $u$ by the corresponding variant of the User-CF algorithm. 
The results, depicted in Figure \ref{fig:intersect}, show that the two-class recommender performs better, 
as expected, since it obtains significantly more ratings. 
Specifically, the expected number of ratings it obtains is almost twice the expected number of ratings the one-class User-CF algorithm obtains. 
After having recommend most of the likable items, mostly non-likable are left to recommend, which explains the inverse $U$-shape in Figure~\ref{fig:intersect}. 

\begin{figure}
\vskip 0.2in
\begin{center}
\begin{tikzpicture}[scale=1]
\begin{groupplot}[group style={group size=1 by 1,horizontal sep=1.4cm,vertical sep=1.2cm,xlabels at=edge bottom, xticklabels at=edge bottom,},
width=0.44\textwidth, height=0.3\textwidth, /tikz/font=\small,
ylabel={$\text{acc-reward}(T)$},
 ylabel shift = -0.15cm
]
\nextgroupplot[xlabel={$T$},
every axis legend/.append style={
        at={(0.3,0.05)},
        anchor=south west}
]
\addplot +[mark=none,solid] table[x index=0,y index=2]{./OneClassKNN/results/comp_onetwo_cpf500x500.dat};
\addlegendentry{one-class}
\addplot +[mark=none,solid] table[x index=0,y index=4]{./OneClassKNN/results/comp_onetwo_cpf500x500.dat};
\addlegendentry{two-class}
\end{groupplot}
\end{tikzpicture}
\end{center}
\vskip -0.2in
\caption{\label{fig:intersect}
Comparison of one and two class recommenders. 
}
\end{figure}
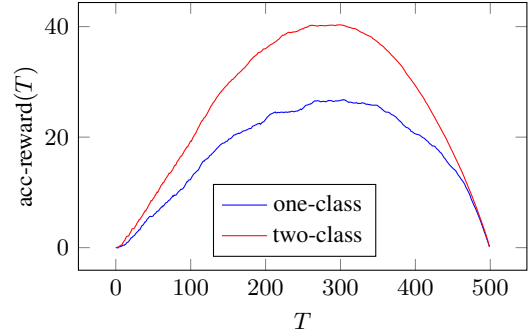

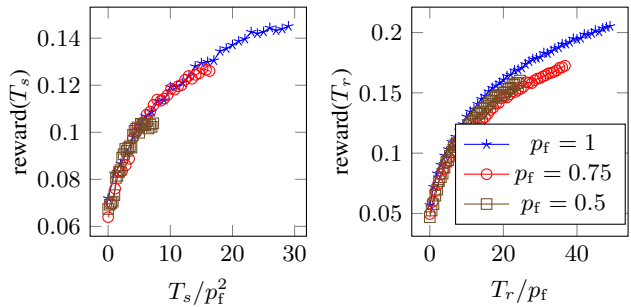
\begin{figure}
\vskip 0.2in
\begin{center}
\begin{tikzpicture}[scale=1]
\begin{groupplot}[group style={group size=2 by 1,horizontal sep=1.4cm,vertical sep=1.2cm,xlabels at=edge bottom, xticklabels at=edge bottom,},
width=0.26\textwidth, height=0.27\textwidth, /tikz/font=\small,
ylabel={$\text{acc-reward}(T)$},
 ylabel shift = -0.15cm
]

\nextgroupplot[xlabel={$T_s/\pf^2$}, ylabel shift = -0.15cm, ylabel = {reward($T_s$)},
every axis legend/.append style={
        at={(0,1)},
        anchor=south west},
        yticklabel style={/pgf/number format/fixed, /pgf/number
      format/precision=3},
]


\addplot +[mark=star] table[x index=0,y index=3]{./OneClassKNN/results/vary_pf_fTj_MLdebiased74_40_fix.dat};
 \addplot +[mark=o] table[x index=6,y index=4]{./OneClassKNN/results/vary_pf_fTj_MLdebiased74_40_fix.dat};
 \addplot +[mark=square] table[x index=7,y index=5]{./OneClassKNN/results/vary_pf_fTj_MLdebiased74_40_fix.dat};

\nextgroupplot[xlabel={$T_r/\pf$},ylabel = {reward($T_r$)},
every axis legend/.append style={
        at={(0.2,0.05)},
        anchor=south west},
    yticklabel style={/pgf/number format/fixed, /pgf/number
      format/precision=3}, ylabel shift = -0.15cm      
]


\addplot +[mark=star] table[x index=0,y index=3]{./OneClassKNN/results/vary_pf_Tr_MLdebiased.dat};
 \addlegendentry{$\pf = 1$}
 \addplot +[mark=o] table[x index=1,y index=4]{./OneClassKNN/results/vary_pf_Tr_MLdebiased.dat};
 \addlegendentry{$\pf = 0.75$}
 \addplot +[mark=square] table[x index=2,y index=5]{./OneClassKNN/results/vary_pf_Tr_MLdebiased.dat};
 \addlegendentry{$\pf = 0.5$}

\end{groupplot}
\end{tikzpicture}
\end{center}
\vskip -0.2in
\caption{\label{fig:intersect} \label{fig:varyTj}
Left: Reward obtained from a single exploitation step, 
 after performing $T_s$ similarity exploration steps and a fixed number of preference exploration steps. 
Right: Reward obtained from a single exploration step, after performing $T_r$ preference exploration steps and a fixed number of similarity exploration steps. 
}
\end{figure}


\paragraph{Dependence of User-CF on $\pf$:}
We next validate empirically that the cold start time and number of  preference exploration steps (needed to learn the preferences of the users) scale as $1/\pf^2$ and $1/\pf$, respectively. 
We start with the former. 
To this end, we split the items into two random disjoint sets $\mc I_1\subset [M]$ and $\mc I_2 \subset [M]$ of equal cardinality. 
We then perform the following experiment for $\pf \in \{ 1,0.75,0.5\}$. 
We start by recommending $\frac{3 M}{k \pf}$ items, chosen uniformly at random from $\mc I_2$ to each user, and, provided the corresponding rating is positive ($[\RML]_{ui} = 1$), we provide this rating to the User-CF algorithm with probability $\pf$. 
The expected number of \emph{positive} ratings 
obtained is therefore independent of $\pf$. 
Those preference exploration steps make sure that the preferences of the items in $\mc I_2$ are explored well. 
We then perform $T_s$ similarity exploration steps on the items in the sets $\mc I_1 = \{i_0,\ldots,i_{M/2-1}\}$, by recommending item $i_{t}$ to user $u$ at $t=0,\ldots,T_s-1$. 
If $[\RML]_{u i_t} = 1$ then we provide the rating $[\RML]_{u i_t}$ to the User-CF algorithm with probability $\pf$. 
After $T_s$ such similarity exploration steps, we perform an exploitation step. 
In Figure \ref{fig:varyTj} we plot the reward defined in \eqref{eq:defrewardnr} obtained by the exploitation step, over $T_s/\pf^2$. 
The results confirm that the cold start time required to find `good' neighborhoods scales inversely proportional to $\pf^2$, since all three curves lie on top of each other. 


Next, we demonstrate that the number of preference exploration steps required to learn the preferences of the users is proportional to $1/\pf$. 
To this end, we perform the same experiment as above, this time, however, we first perform $T_s = 25/\pf^2$ similarity exploration steps on the items in the set $\mc I_1$, and then perform $T_r$ preference exploration steps by recommending $T_r$ items, chosen uniformly at random from $\mc I_2$ to each user. As before, if $[\RML]_{ui} = 1$, the rating $[\RML]_{ui}$ is provided to the algorithm with probability $\pf$. 
In Figure \ref{fig:varyTj} we plot the reward obtained from performing a single exploitation step after $T_r$ such preference exploration steps over $T_r/\pf$. 
The results indicate that, as predicted by our theory, the number of preference exploration steps required to learn the preferences is proportional to $\pf$, as the curves for different $\pf$ lie on top of each other.  
As mentioned previously, this is not surprising, as the number of positive ratings obtained is proportional to $\pf$.

\bibliography{../computerscience.bib}
\bibliographystyle{icml2017}

\onecolumn

\newpage 

\section*{Supplementary Material for
``The Sample Complexity of Online One-Class Collaborative Filtering''}

\section{\label{sec:proof}Proof of Theorem \ref{thm:mainspecial}}

Theorem~\ref{thm:mainspecial} follows immediately from the following result. 

\begin{theorem}
\label{thm:UBmaingen}
Suppose that there are at least $\frac{N}{2K}$ users of the same type, for all user types, 
and assume that at least a fraction $\nu$ of all items is likable to a given user, for all users.  
Moreover, suppose that for some $\gamma \in [0,1)$, all users satisfy condition~\eqref{eq:userssuffdiffthm}. 
Pick $\delta >0$ and suppose that the number of nearest neighbors $k$, the batch size $Q$, and the parameter $\eta$, are chosen such that $k \leq \frac{9 N}{40K}$, $\eta \leq \nu/2$,  
\begin{align}
\frac{k}{Q}  \geq  \frac{64 \log( 8M / \delta)}{\pf \Delta^2 } ,
\label{eq:batchthm}
\end{align}
and 
\begin{align}
Q \geq \frac{10}{\nu} \log(4/\delta).
\label{eq:Qgeqinus}
\end{align}
Then the reward accumulated by the 
User-CF algorithm up to time $T \in [T_{\text{start}} , \frac{4}{5} \nu \nitem \pf]$ with 
\[
T_{\text{start}}
=
\frac{
\left(  512 \max\left(\log\left( \frac{4NQ}{k\Delta} 
\right), \log\left(\frac{88}{\delta}\right)\right) 
\right)^{\frac{1}{1-\alpha}}
}{
(3  \pf^2  (1 - \gamma)^2 \nu)^{\frac{1}{1-\alpha}}
\left(1 - \max\left( \frac{1}{T} , \frac{2}{\eta Q}\right)\right) 
}
%
%
\]
satisfies
\begin{align}
\frac{\arewardT}{NT}  \geq \left( 1 - \frac{T_{\text{start}} }{T} 
- 2^\alpha \frac{ (T - T_{\text{start}})^{1-\alpha} }{T(1-\alpha)} -  \max \left( \frac{1}{T} , \frac{2}{\eta Q} \right) \right) (1-\delta). 
\label{eq:lbonregret}
\end{align}
\end{theorem}

\indent Theorem~\ref{thm:mainspecial} follows by choosing the parameter of the User-CF algorithm as follows:
\begin{align*}
\eta = \frac{\nu}{2}, 
\quad
k = \frac{9}{40} \frac{N}{K},
\quad 
\text{and}
\quad 
Q =  k 
\frac{\pf \Delta^2 }{64 \log( 8M / \delta)}.
\end{align*}
To see this, note that by definition, the conditions on $k$ and $\eta$ and condition~\eqref{eq:batchthm} on $Q$ are satisfied. 
By~\eqref{eq:kappacond}, condition~\eqref{eq:Qgeqinus} holds and 
$\frac{2}{\eta Q} = \frac{K}{N} \frac{c'\log( M / \delta )}{\pf \Delta^2}$.
Moreover, 
$\max\left(\log\left( \frac{4NQ}{k\Delta} \right), \log \left( \frac{88}{\delta} \right) \right) \leq \tilde c \log(N/\delta)$.


\subsection{\label{sec:proof}Proof of Theorem \ref{thm:UBmaingen}}

Theorem~\ref{thm:UBmaingen} is proven by showing that at time $t \geq T_{\text{start}}$ the following holds for all users $u$: 

\begin{enumerate}[i)]

\item \label{it1} the neighborhood of $u$ is sufficiently well explored by similarity exploration steps so that most of the nearest neighbors of $u$ are \emph{good}, i.e., are of the same user type as $u$ (similarly, neighbors are called \emph{bad} if they are of a different user type than $u$), 

\item \label{it2} for $t\geq T_{\text{start}}$, the estimates $\hat p_{ui}$, for all $i \in \setQ_q, q=0,\ldots,\frac{t}{\eta Q}-1$ correctly predict whether $i$ is likable by $u$ or not, and 

\item \label{it3}  there exist items in the sets $\setQ_q, q=0,\ldots,\frac{t}{\eta Q}-1$ that are likable by $u$ and that have not been rated by $u$ at previous times steps. 

\end{enumerate}
Conditions~\ref{it1}, \ref{it2}, and \ref{it3} guarantee that an exploitation step recommends a likable item.

Formally, we start by defining the following events: 
\begin{align}
\mc G_\beta(t) 
=\{\text{At time $t$, no more than $\beta k$ of the $k$-nearest neighbors of $u$ are bad} \},
\end{align}
%
\begin{align}
\mc L(t) = &\{\text{at time $t$, there exists an item $i \in \setQ_q$}, \nonumber \\
&\hspace{0.5cm}\text{$q =0,\ldots,t/(\eta Q)-1$ that is likable by $u$}\}, 
\label{eq:defLu}
\end{align}
and
\begin{align}
\mc E(t) = \bigcup_{q=0,\ldots,\frac{t}{\eta Q}-1} \mc E_{q}(t),
\end{align}
with
\begin{align}
\mc E_{q}(t)
=
&\{\text{Conditioned on $\mc G_{\frac{\Delta}{4Q}}(t)$, for all $i\in \setQ_q$, } \nonumber \\
&\hat p_{ui} > \pf/2, \text{ if } p_{ui} > 1/2 + \Delta,\text{ and} \nonumber \\
&\hat p_{ui} < \pf/2, \text{ if } p_{ui} < 1/2 - \Delta
\}.
\label{def:Eui} 
\end{align}
For convenience, we omit in the notion of $\mc L(t)$, $\mc G_\beta(t)$, $\mc E(t)$, and $\mc E_{q}(t)$ the dependence on $u$.   
The significance of those definitions is that if  $\mc L(t)$, $\mc G_{\frac{\Delta}{4Q}}(t)$, and $\mc E(t)$ hold simultaneously, then the recommendation made to user $u$ by an exploitation step at time $t$ is likable. 
We can therefore lower-bound the reward $\arewardT$ as follows: 
\begin{align}
\frac{\arewardT}{NT} 
%
&=  \frac{1}{NT} \sum_{t=0}^{T-1} \sum_{u=0}^{N-1} \PR{X_{u i(u,t)}=1} 
\nonumber \\
&\geq 
\frac{1}{NT} 
\sum_{u=0}^{N-1}
\sum_{t=0, t \notin \{ \eta Q q\colon q=0,1,\ldots \} }^{T-1} 
\PR{ \text{exploitation at $t$} } \PR{X_{u i(u,t)}=1 | \text{exploitation at $t$}} \label{eq:independence} \\
&\geq
\frac{1}{N}  \sum_{u=0}^{N-1}
\left(
\frac{1}{T}
\sum_{t=0}^{T-1}
(1 - (2/t)^{\alpha}) \PR{X_{u i(u,t)}=1 | \text{exploitation at $t$}}  
- 
\max \left( \frac{1}{T} , \frac{2}{\eta Q} \right)
\right)
\label{eq:useubrexpsteps}\\
&\geq
\frac{1}{N}  \sum_{u=0}^{N-1}
\left(
\frac{1}{T}
\sum_{t=T_{\text{start}}}^{T-1}
(1-\delta) (1 - (2/t)^{\alpha} )
- 
\max \left( \frac{1}{T} , \frac{2}{\eta Q} \right)
\right)
\label{eq:usebfexpexpt} \\
&\geq (1-\delta) \left( 1 - \frac{T_{\text{start}} }{T} 
- 2^\alpha \frac{ (T - T_{\text{start}})^{1-\alpha} }{T(1-\alpha)} 
- \max \left( \frac{1}{T} , \frac{2}{\eta Q} \right)
\right) \label{eq:uselbsum}.
\end{align}
Here, \eqref{eq:independence} follows from 
\[
\PR{X_{u i(u,t)}=1| \text{preference exploration at $t$} } \geq 0 
\quad \text{and} \quad
\PR{X_{u i(u,t)}=1| \text{similarity exploration at $t$} } \geq 0. 
\]
For
\eqref{eq:useubrexpsteps} we used, for $t\neq \eta Q q$, 
\[
\PR{\text{exploration at $t$}} 
= 1-(t- \lfloor t /(\eta Q)\rfloor )^{-\alpha}
\geq 1 - (t(1 - 1 /(\eta Q)))^{-\alpha}
\geq 1 - (2/t)^\alpha
\] 
which follows from $\eta Q \geq 2$. 
Moreover we used for \eqref{eq:useubrexpsteps} that the fraction of preference exploration steps up to time  $T$ is at most $\max( \frac{1}{T} , \frac{2}{\eta Q})$. 
To see that, note that at $T \in \{\eta Qq,\ldots,  \eta Q(q+1)\}$ we have performed $q+1$ preference exploration steps. 
It follows that, for $q\geq 1$, the fraction of preference exploration steps performed up to $T$ is given by
$\frac{q+1}{q \eta Q} \leq \frac{2}{\eta Q}$. 
Thus, for any $T\geq 1$, the fraction of preference exploration steps is $\leq \max( \frac{1}{T} , \frac{2}{\eta Q})$. 
Equality~\eqref{eq:usebfexpexpt} follows from 
\begin{align}
\PR{X_{u i(u,t)}=1 | \text{exploitation at $t$}}
&\geq 
\PR{\mc E(t) \cap \mc G_{\frac{\Delta}{4Q}}(t) \cap  \mc L(t)} \nonumber \\
&\geq 1-\delta. 
\label{eq:probEutge1de}
\end{align}
Here, inequality~\eqref{eq:probEutge1de} holds for $t \geq T_{\text{start}}$ and is established below. 
Finally, inequality~\eqref{eq:uselbsum} follows from 
\begin{align}
\sum_{t=T_{\text{start}}}^{T-1}  
t^{-\alpha}
&\leq  \int_{T_{\text{start}} -1}^{T-1} t^{-\alpha}
= \frac{1}{1-\alpha} t^{1-\alpha} |_{t=T_{\text{start}} -1}^{T-1} \nonumber \\
&= \frac{ (T-1)^{1-\alpha} - (T_{\text{start}}-1)^{1-\alpha} }{1-\alpha} 
\leq 
\frac{ (T - T_{\text{start}})^{1-\alpha} }{1-\alpha}. \nonumber
\end{align}
It remains to establish \eqref{eq:probEutge1de}. 
To this end, define for notational convenience 
\[
A 
\defeq 
\frac{256 \max\left(\log\left( \frac{4NQ}{k\Delta} \right), \log\left(\frac{88}{\delta}\right)\right) }{ 3  \pf^2  (1 - \gamma)^2 \nu }, 
\]
and let $T_s$ be the number of similarity exploration steps executed up to time $T$. 
Inequality~\eqref{eq:probEutge1de} follows by noting that, for all $t\geq T_{\text{start}}$, by the union bound, 
\begin{align}
\PR{\comp{(\mc E(t) \cap \mc G_{\frac{\Delta}{4Q}}(t) \cap \mc L(t) )} } 
&\leq 
\PR{\comp{\mc E}(t) } 
+
\PR{\comp{\mc G}_{\frac{\Delta}{4Q}}(t)}
+
\PR{\comp{\mc L}(t) } \nonumber \\
&\leq 
\PR{ \comp{\mc E}(t)}
+ 
\PR{ \comp{\mc G}_{\frac{\Delta}{4Q}}(t) | T_s \geq A } +
\PR{ T_s \leq A }
+
\PR{\comp{\mc L}(t) } \label{eq:evsplit} \\
&\leq 
\frac{\delta}{4} + \frac{\delta}{4} + \frac{\delta}{4} + \frac{\delta}{4} = \delta. 
\label{eq:importantbounds}
\end{align}
Here, inequality~\eqref{eq:evsplit} follows since for two events $C,B$ we have that
\begin{align}
\label{eq:events}
\PR{C}
=
\PR{C \cap B} + \PR{C \cap \comp{B} }
=
\PR{C|B} \PR{B} + \PR{C|\comp{B}} \PR{ \comp{B} }
\leq 
\PR{C | B} + \PR{\comp{B}}.
\end{align}
Inequality~\eqref{eq:importantbounds} follows from 
\begin{align}
\PR{ \comp{\mc E}(t)  } \leq \delta/4 \label{eq:compEulde} \\
\PR{ \comp{\mc G}_{\frac{\Delta}{4Q}}(t) |  T_s \geq A  } \leq \delta/4 \label{eq:neighborhooddel4}\\
\PR{  T_s \leq A } \leq \delta/4 \label{eq:PrTjA}\\
\PR{\comp{\mc L}(t) } \leq \delta/4 \label{eq:PrLut}.   
\end{align}
In the remainder of this proof, we establish the inequalities~\eqref{eq:compEulde}-\eqref{eq:PrLut}. 
The key ingredient for these bounds are concentration inequalities, in particular a version of Bernstein's inequality \cite{bardenet_concentration_2015}. 

\paragraph{Proof of \eqref{eq:compEulde}:}
By the union bound, we have, for all $t=0,\ldots,M-1$, that 
\[
\PR{\comp{\mc E}(t) } 
%
%
\leq \sum_{q=0}^{M/Q-1}
\PR{\comp{\mc E}_{q}(t) }
\leq \frac{\delta}{4}
\]
as desired. 
Here, we used $\PR{\comp{\mc E}_{q}(t) } \leq \frac{\delta Q}{4M}$, which follows from Lemma \ref{lem:randexpbatch} stated below with $\delta'= \frac{\delta Q}{4M}$ and $T_r=1$ (note that the assumption \eqref{eq:batch} of Lemma \ref{lem:randexpbatch} is implied by the assumption \eqref{eq:batchthm} of Theorem \ref{thm:UBmaingen}). 

\begin{lemma}[Preference exploration]
Suppose we recommend $T_r$ random items to each user, chosen uniformly at random from a set $\setQ \subseteq [M]$ of $Q$ items. 
Suppose that $p_{vi}$ is $\Delta$-bounded away from $1/2$, for all $i \in \setQ$ and for all $v \in \mc N_u$, where $\mc N_u$ is a set of $k$ users, of which no more than $\beta k$, with $\beta \leq \frac{\Delta T_r}{4Q}$, of the users are of a different type than $u$. 
Fix $\delta'>0$. If 
\begin{align}
T_r  \frac{k}{Q} \frac{\pf \Delta^2 }{ 64 \log(2 Q / \delta')} \geq 1
\label{eq:batch}
\end{align}
then, with probability at least $1- \delta'$, for all $i \in \setQ$, 
$\hat p_{ui} > \frac{\pf}{2}$ if $p_{ui} \geq 1/2 + \Delta$ and $\hat p_{ui} < \frac{\pf}{2}$ if $p_{ui} \leq 1/2 - \Delta$. 
\label{lem:randexpbatch}
\end{lemma}

\paragraph{Proof of \eqref{eq:neighborhooddel4}:}
Inequality~\eqref{eq:neighborhooddel4} follows from Lemma \ref{lem:mgfb} below, which ensures that a user has many good and only few bad neighbors. 
\begin{lemma}[Many good and few bad neighbors]
Let $\mc T_u$ be the subsets of all users $[N]$ that are of the same type of $u$ and suppose its cardinality satisfies $\geq \frac{N}{2K}$. 
Suppose that, for some constant $\gamma \in [0,1)$, 
condition~\eqref{eq:userssuffdiffthm} holds, and that the number of nearest neighbors $k$ satisfies $k \leq   \frac{9N}{40K}$. 
Choose $\beta \in (0,1)$, 
and suppose 
\begin{align}
 T_s  
\geq 
\frac{64 \log(N/(\beta k ))}{ 3  \pf^2  (1 - \gamma)^2 \frac{1}{M} \min_{v\in \mc T_u} \innerprod{\vp_u}{\vp_v} }
\label{eq:TlargeK}
\end{align}
similarity exploration steps have been performed. 
Then, with probability at least
$1 - 11 e^{ - \frac{3}{ 64 }  T_s  \pf^2 (1-\gamma)^2  \frac{1}{M}  \min_{v\in \mc T_u} \innerprod{\vp_u}{\vp_v} }$,
the set of nearest neighbors $\mc N_u$ of user $u$ (defined in Section \ref{sec:algformal}), contains no more than $\beta k$ bad neighbors. 
\label{lem:mgfb}
\end{lemma}
To see that inequality~\eqref{eq:neighborhooddel4} follows from Lemma \ref{lem:mgfb}, we first note that $T_s \geq A$ guarantees that condition~\eqref{eq:TlargeK} of Lemma \ref{lem:mgfb} is satisfied (with $\beta = \frac{\Delta}{4Q}$). To see this, note that since each user likes at least a fraction $\nu$ of the items, we have
\begin{align}
\frac{1}{M} \min_{v\in \mc T_u} \innerprod{\vp_u}{\vp_v}
\geq \nu \left( \frac{1}{2} + \Delta \right)^2 \geq \frac{\nu}{4}. 
\label{eq:iqebminlb}
\end{align}
Lemma \ref{lem:mgfb} therefore implies 
\[
\PR{ \comp{\mc G}_{\frac{\Delta}{4Q}}(t) | T_s \geq A }
\leq 
11 e^{ - \frac{3}{ 64 }  T_s  \pf^2 (1-\gamma)^2  \frac{1}{M}  \min_{v\in \mc T_u} \innerprod{\vp_u}{\vp_v} }
\leq
 11 e^{- \log(88/\delta)}
= \frac{\delta}{8},
\]
as desired. 
For the second inequality above we used \eqref{eq:iqebminlb} and $T_s \geq A$.



\paragraph{Proof of \eqref{eq:PrTjA}:}
We next establish the inequality $\PR{T_s \leq A} \leq \delta/4$. 
To this end, recall that a similarity exploration step is carried out at $t = 0,\ldots, T-1, t \neq \eta Q q, q = 0,1,\ldots$ with probability $1/ (t - \lfloor t/(\eta Q) \rfloor)$. 
Recall from the discussion below inequality~\eqref{eq:uselbsum}, 
that the fraction of time steps up to time $T$ for which $t = \eta Q q$, for some $q$, is at most $\max( \frac{1}{T} , \frac{2}{\eta Q})$. 
It follows that the number of similarity exploration steps, $T_s$, carried out after $t \geq T_{\text{start}}$ steps of the User-CF algorithm, stochastically dominates 
the random variable 
$S = \sum_{t=1}^{\tilde T} Z_t$, $\tilde T = T_{\text{start}} (1 - \max( \frac{1}{T} , \frac{2}{\eta Q}) )$,  where $Z_t$ is a binary random variable with $\PR{Z_t =1} = 1/t^\alpha$. 
It follows that 
\begin{align}
\PR{T_s \leq A} 
=
\PR{T_s \leq  \tilde T^{1-\alpha}/2  }
\leq 
e^{- \frac{ \tilde T^{1-\alpha} }{20}}
%
\leq
\delta/4, 
\label{eq:probTjA}
\end{align}
where the first inequality holds by definition of $T_{\text{start}}$, i.e., 
\[
T_{\text{start}}
=
(2A)^{\frac{1}{1-\alpha}} / \left(1 - \max\left( \frac{1}{T} , \frac{2}{\eta Q}\right) \right), 
\]
and the second inequality holds by Lemma \ref{lem:sumrvdec} stated below. Finally, the last inequality in \eqref{eq:probTjA} follows from 
\[
\tilde T = (2A)^{\frac{1}{1-\alpha}} \geq \frac{128}{3} \log(44/\delta). 
\]

The following lemma appears in \cite{bresler_latent_2014}. 
\begin{lemma}
Let $S = \sum_{t=1}^{\tilde T} Z_t$ where $Z_t$ is a binary random variable with $\PR{Z_t =1} = 1/t^\alpha$, $\alpha \in (0,4/7)$. 
We have that 
\[
\PR{
S_T \leq \tilde T^{1-\alpha}/2
}
\leq 
e^{- \frac{ \tilde T^{1-\alpha}}{20}}.
\]
\label{lem:sumrvdec}
\end{lemma}

\paragraph{Proof of \eqref{eq:PrLut}:}
Suppose $t < \eta Q$, 
consider user $u$, and let $N_0$ be the total number of  items likable by $u$ in the set $\setQ_0$ (recall that $\setQ_0$ is choosen uniformly at random from the subset of items $[M]$ of cardinality $Q$). 
Note that $N_0 > \eta Q$ implies that at $t < \eta Q$, there exist items that are likable by $u$ in $\setQ_0$ that have not been recommended to $u$ yet. 
Therefore, we can upper bound the probability that no likable items are left to recommend, for $t < \eta Q$, by
\begin{align}
\PR{ \comp{\mc L}(t) }
\leq
\PR{N_0 \leq \eta Q}
&\leq
\PR{ N_0 \leq Q \nu/2 }
\leq
\PR{N_0 \leq \EX{N_0} - Q \nu/2 } \label{eq:useglnu} \\
&\leq 
e^{-Q \frac{ (\nu/2)^2}{2 \nu (1-\nu) + \frac{2}{3} \frac{\nu}{2}}
}
=
e^{-Q \frac{ \nu/4}{2 (1-\nu) + \frac{1}{3}}}
\leq 
e^{-Q \frac{\nu}{10}}
\leq \frac{\delta}{4}.
\label{eq:uBsonQset}
\end{align}
Here, the first inequality in \eqref{eq:useglnu} follows from $\eta \leq \nu/2$, by assumption; the second inequality in \eqref{eq:useglnu} follows from $\EX{N_0} \geq \nu Q$ (since at least a fraction of $\nu$ of the items is likable by $u$), 
the first inequality in \eqref{eq:uBsonQset} follows from Bernstein's inequality \cite{bardenet_concentration_2015}, and finally the last inequality in \eqref{eq:useglnu} holds by assumption \eqref{eq:Qgeqinus}. 
We have established that $\PR{\comp{\mc L}(t) } \leq \delta/4$, for $t < \eta Q$. 
Using the exact same line of arguments yields the same bound for $t \in [\eta Q,  \eta M]$. 

It remains to upper bound $\PR{\comp{\mc L}(t)}$ for $t\in [\eta M , \frac{4}{5} \nu M \pf]$. 
To this end, let $N_{u}^c(T)$ be the number of (likable) items that have been 
rated by user $u$ after $T$ time steps, and note that 
if $N_{u}^c(T)$ is strictly smaller than the (minimum) number of likable items, then there are likable items left to recommend. 
Formally, 
\begin{align}
\PR{ \comp{\mc L}(t) }
\leq 
\PR{ N_u^c(T) \geq \nu M}
\label{eq:Nucgemu1}
\end{align}
where we used that for each user $u$, at least $\nu M$ items are likable. 
Recall that with probability $p_{ui}\pf \leq \pf$ a likable item $i$ is rated if it is recommended to $u$. Once rated, an item is not recommended again. 

Note that $N_{u}^c(T)$ is statistically dominated by a sum of independent binary random variables $Z_t$ with $\PR{Z_t = 1} = \pf$. 
We therefore have that 
\begin{align}
\PR{ N_u^c(T) \geq \nu M }
\leq 
\PR{ N_{u}^c(T) \geq T (\pf +  \frac{\pf}{4})  }  
\leq e^{  -\frac{T \pf^2}{2} }
\leq e^{  -\frac{T_{\text{start}} \pf^2}{2} }
\leq \frac{\delta}{4}.
\label{eq:Nucgemu2}
\end{align}
Here, the first inequality holds by the assumption $T \leq \frac{4}{5} \nu M \pf$, the second inequality follows by Hoeffding's inequality, the third inequality follows by $T\geq T_{\text{start}}$, and the last inequality follows from 
$T_{\text{start}}  \geq \frac{2}{\pf^2} \log(4/\delta)$, which holds by definition of $T_{\text{start}}$. 
Application of \eqref{eq:Nucgemu2} on \eqref{eq:Nucgemu1} concludes the proof of $\PR{\comp{\mc L}(t) } \leq \delta/4$. 




\subsection{Proof of Lemma \ref{lem:mgfb} }

\newcommand{\perrg}{p_{\mathrm{good}}}
\newcommand{\perrb}{p_{\mathrm{bad}}}

Recall that $\vratingS_{u} \in \{0,1\}^M$ is the  vector containing the responses $\ocR_{ui}$ of user $u$ to previous \emph{similarity} exploration steps up to time $t$, and that we assume in Lemma \ref{lem:mgfb}, that $T_s$ similarity exploration steps have been performed up to time $t$. 
To establish Lemma \ref{lem:mgfb}, we show that there are more than $k$ users $v$ that are of the same user type as $u$ and satisfy
$
\frac{1}{T_s}\innerprod{\vratingS_u}{\vratingS_v} \geq \theta $, 
and at the same time, there are fewer than $k\beta$ users of a different user type as $u$ that satisfy 
$
\frac{1}{T_s}\innerprod{\vratingS_u}{\vratingS_v} \geq \theta
$
for a certain threshold $\theta$ chosen below. 
This is accomplished by the following two lemmas.

\begin{lemma}[Many good neighbors]
Suppose there are at least $\frac{N}{2K}$ users of the type as user $u$ (including $u$), and suppose 
that $T_s$ similarity exploration steps have been performed. 
Then, with probability
at least $1-10\perrg$, 
\[
\perrg \defeq 
e^{  - \frac{3}{ 16 }  T_s  p_g(1-\theta/p_g)^2  },  \quad p_g \defeq \pf^2 \frac{1}{M}  \min_{v\in \mc T_u} \innerprod{\vp_u}{\vp_v}, 
\]
at least $\frac{9 N}{40 K}$ users $v$ of the same user type as $u$ obey $\frac{1}{T_s} \innerprod{\vratingS_u }{\vratingS_v} \geq \theta$. 
\label{lem:manygood}
\end{lemma}
\begin{lemma}[Few bad neighbors]
Suppose that $T_s$ similarity exploration steps have been performed. 
Then, with probability at least $1 -  \perrb$, where 
\[
\perrb = 
e^{ - 
\frac{T_s p_b (\theta/p_b - 1)^2 /4}{ 1 + (\theta/p_b - 1)/3 } 
}, \quad 
p_b \defeq \pf^2 \max_{v \notin \mc T_u} \frac{1}{M} \innerprod{\vp_v}{\vp_u}, 
\]
at most $N \perrb$ users $v$ of a different user type than $u$ obey $\frac{1}{T_s} \innerprod{\vratingS_u }{\vratingS_v} \geq \theta $. 
\label{lem:fewbad}
\end{lemma}

We set 
\[
\theta = \frac{p_g + p_b}{2}.
\]
With this choice, by Lemma \ref{lem:manygood}, there are more than $\frac{9N}{40K} \geq k$ (the inequality holds by assumption) 
users $v$ of the same type as $u$ that satisfy 
$
\frac{1}{T_s}\innerprod{\vratingS_u}{\vratingS_v} \geq \theta
$, with probability at least $1- 10\perrg$. 
By Lemma \ref{lem:fewbad}, there are no more than $N\perrb$ users $v$ of a different type as $u$ with $
\frac{1}{T_s}\innerprod{\vratingS_u}{\vratingS_v} \geq \theta
$. 
Thus, by the union bound, $\mc N_u$ contains less than $\perrb N$ bad neighbors with probability at least 
\[
1- 10\perrg - \perrb \geq 1 - 11 e^{ - \frac{3}{ 64 }  T_s  p_g(1-\gamma)^2 }.
\]
Here, we used 
\[
\perrg
=
e^{ - \frac{3}{ 64 }  T_s  p_g(1-p_b/p_g)^2 }
\leq e^{ - \frac{3}{ 64 }  T_s  p_g(1-\gamma)^2 }
\]
where the inequality follows by $p_b/p_g \leq \gamma$, by \eqref{eq:userssuffdiffthm}. 
Moreover, we used 
\begin{align}
\perrb 
&= 
e^{ - 
\frac{T_s p_b ( \theta/p_b - 1)^2 /4}{ 1 + (\theta/p_b - 1)/3 } 
}
=
e^{ - 
\frac{T_s p_b ( p_g/p_b - 1)^2 /16}{ 1 + (p_g/p_b - 1)/6 } 
}
=
e^{ - 
\frac{T_s p_g ( \sqrt{p_g/p_b} - \sqrt{p_b/p_g} )^2 /16}{ 1 + (p_g/p_b - 1)/6 } 
}
\leq
e^{ - 
\frac{T_s p_g ( \sqrt{1/\gamma} - \sqrt{\gamma} )^2 /16}{ 1 + (1/\gamma - 1)/6 } 
} \nonumber \\
&\leq
e^{ - 
\frac{T_s p_g ( \sqrt{1/\gamma} - \sqrt{\gamma} )^2 /16}{ 1 + (1/\gamma - 1) }
= 
e^{ - 
T_s p_g ( 1 - \gamma )^2 /16 } 
}. \label{eq:qdbound}
\end{align}
Here, the first inequality follows from the absolute value of the exponent being decreasing in $p_b/p_g$, and from the assumption $p_b/p_g \leq \gamma$, by \eqref{eq:userssuffdiffthm}. 

To conclude the proof, we needed to establish that the maximum number of bad neighbors $N  \perrb$ satisfies $N \perrb \leq \beta k$. This follows directly by noting that, by assumption \eqref{eq:TlargeK}, the RHS of \eqref{eq:qdbound} is upper-bounded by $\frac{\beta k}{N}$. 


\subsubsection{Proof of Lemma \ref{lem:manygood}}

Consider $u$ and assume there are exactly $\frac{N}{2K}$ users from the same user type. There could be more, but it is sufficient to consider $\frac{N}{2K}$. 
Let $v$ be of the same user type. 
We start by showing that $\frac{1}{T_s}\innerprod{\vratingS_u }{\vratingS_v} \geq \theta$ with high probability. 
To this end, note that $\innerprod{\vratingS_u }{\vratingS_v}
= \sum_{t=0}^{T_s-1} 
\ocR_{u \pi(t)} \ocR_{v \pi(t)}
$ where $\pi$ is the random permutation of the item space 
drawn by the User-CF algorithm at initialization, 
and $\ocR_{u \pi(t)} \ocR_{v \pi(t)}$ is a binary random variable, independent across $t$, with success probability $\pf^2 p_{u \pi(t)} p_{v \pi(t)}$. 
Setting $a \defeq \pf^2 \frac{1}{M} \innerprod{\vp_u}{\vp_v}$, for notational convenience, 
it follows that 
\begin{align}
\PR{ \frac{1}{T_s} \innerprod{\vratingS_u }{\vratingS_v}  
\leq \theta }
&=
\PR{ \frac{1}{T_s} \innerprod{\vratingS_u }{\vratingS_v}  \leq a  - (a - \theta)} \\
&\leq 
e^{ - 
\frac{T_s (a - \theta)^2 /2}{ a + (a - \theta)/3 } 
}  \label{eq:usebinruv} \\
&=
e^{ - 
\frac{T_s a(1 - \theta/a)^2 /2}{ 1 + (1 - \theta/a)/3 } 
}
\leq e^{  - \frac{3}{ 8 } T_s a  ( 1 - \theta/ a  )^2}  \label{eq:padflethem1} \\
&\leq e^{  - \frac{3}{ 8 } T_s  p_g  ( 1 - \theta/ p_g  )^2 } \leq \perrg. 
\label{eq:padflethe}
\end{align}
Here, \eqref{eq:usebinruv} follows from Bernstein's inequality \cite{bardenet_concentration_2015}, and for \eqref{eq:padflethe} we used that the RHS of \eqref{eq:padflethem1} is decreasing in $a$. 

Next, consider the random variable 
\[
W 
= \sum_{v \in \mc T_u} G_v,
\quad G_v  
= \ind{ \frac{1}{T_s} \innerprod{\vratingS_u }{\vratingS_v}  \geq \theta },
\]
where $\mc T_u$ is the subset of all users $[N]$ that are of the same time as user $u$, as before. By Chebyshev's inequality, 
\begin{align}
\PR{W - \EX{ W } \leq - \frac{\EX{W}}{2} }
\leq 
\frac{\mathrm{Var}(W)}{ ( \EX{W}/2 )^2 }.
\label{eq:cheb}
\end{align}
Since there are at least $\frac{N}{2K}$ users of the same type, the carnality of $\mc T_u$ is lower bounded by $\frac{N}{2K}-1$. It follows with \eqref{eq:padflethe} that 
\[
\EX{W} \geq (1-\perrg) \left(\frac{N}{2K}-1\right).
\]
Next, we upper bound the variance of $W$. We have
\[
\mathrm{Var}(W) = \sum_{v\in \mc T_u} \mathrm{Var}(G_v) + \sum_{v, w \in \mc T_u, v\neq w} \mathrm{Cov}(G_v, G_w).
\]
With $G_v = G_v^2$, 
\[
\mathrm{Var}(G_v) = \EX{G_v^2} - \EX{G_v}^2 = \EX{G_v}(1 - \EX{G_v}) \leq 1 - \EX{G_v} \leq \perrg. 
\]
Similarly, 
\[
\mathrm{Cov}(G_v,G_w) = \EX{G_v G_w} - \EX{G_v}\EX{G_w} \leq 1 - (1-q)^2 \leq 2 \perrg.
\]
Thus, we obtain 
\[
\mathrm{Var}(W) \leq \left( \frac{N}{2K} - 1 \right) \perrg +  \left( \frac{N}{2K} -1 \right)  \left( \frac{N}{2K} -2 \right) 2 \perrg
\leq 
\left( \frac{N}{2K} - 1 \right)^2 2 \perrg. 
\]
Plugging this into \eqref{eq:cheb} yields 
\[
\PR{W - \EX{ W } \leq - \frac{\EX{W}}{2} }
\leq 
\frac{8\perrg}{ (1 - \perrg)^2 } \leq 10 \perrg,
\]
for $\perrg \leq 1/10$. 
It follows that the number of good neighbors is larger than
\[
W \geq \EX{W}/2 \geq (1 - \perrg) \frac{N}{4K} \geq \frac{9 N}{40 K}
\]
with probability at least $1 - 10 \perrg$. 


\subsubsection{Proof of Lemma \ref{lem:fewbad}}
Let $u$ and $v$ be two fixed users of different user types. 
Similarly as in the proof of Lemma \ref{lem:manygood}, 
we start by showing that $\frac{1}{T_s}\innerprod{\vratingS_u }{\vratingS_v} \leq \theta$ with high probability. 
To this end, note that $\innerprod{\vratingS_u }{\vratingS_v}
= \sum_{t=0}^{T_s-1} 
\ocR_{u \pi(t)} \ocR_{v \pi(t)}
$ where $\pi$ is a random permutation of the item space and $\ocR_{u \pi(t)} \ocR_{v \pi(t)}$ is a binary random variable, independent across $t$, with success probability $\pf^2 p_{u \pi(t)} p_{v \pi(t)}$. 
Setting $a = \pf^2 \frac{1}{M} \innerprod{\vp_u}{\vp_v}$, for notational convenience, 
it follows that 
\begin{align}
\PR{ \frac{1}{T_s} \innerprod{\vratingS_u }{\vratingS_v}    \geq \theta }
&=\PR{ \frac{1}{T_s} \innerprod{\vratingS_u }{\vratingS_{v'}}    \geq  a  + (\theta - a) } \nonumber \\
&
\leq e^{ - 
\frac{T_s (\theta - a)^2 /2}{ a + (\theta - a)/3 } 
} \label{eq:withep} \\
&\leq e^{ - 
\frac{T_s p_b (\theta/p_b - 1)^2 /2}{ 1 + (\theta/p_b - 1)/3 } 
}  = \perrb^2  . \label{eq:aftbernst}
\end{align}
Here, \eqref{eq:withep} follows from Bernstein's inequality. Specifically, we use that $\pi$ is a random permutation of the item space as well as that $\ocR_{u i} \ocR_{v i}$ are binary random variables independent across $i$ (note that Bernstein's inequality also applies to sampling without replacement, see e.g., \cite{bardenet_concentration_2015}). 
Finally, for inequality~\eqref{eq:aftbernst}, we used that $a\leq p_b = \pf^2 \max_{v \notin \mc T_u} \frac{1}{M} \innerprod{\vp_v}{\vp_u}$. 


Set $N_{bad} = \sum_{v \notin \mc T_u} \ind{\text{$u$ and $v$ are declared neighbors}}$. 
By inequality~\eqref{eq:aftbernst}, we have
$
\EX{N_{\text{bad}}} \leq \perrb^2 N. 
$
Thus, by Markov's inequality, 
\[
\PR{ N_{\text{bad}}  \geq N \perrb } 
\leq \frac{\EX{N_{\text{bad}}} }{ N \perrb }
\leq \frac{\perrb^2 N }{ N \perrb }
= \perrb,
\]
which concludes the proof. 

\subsection{Proof of Lemma \ref{lem:randexpbatch} (preference exploration)}

\newcommand\nb{\beta k}
%
Assume w.l.o.g.~that $p_{ui}>1/2 + \Delta$, for all $i \in \setQ$. The case where some of the $p_{ui}$ satisfy $p_{ui} < 1/2 - \Delta$ is treated analogously. 
To prove Lemma \ref{lem:randexpbatch}, we may further assume that  $p_{ui}  = \frac{1}{2} + \Delta$, for all $i \in \setQ$, since $\PR{\hat p_{ui} > \frac{\pf}{2}}$ is increasing in $p_{ui}$. 

Consider a fixed item $i \in \setQ$, and let $\mc N_u^{\text{good}}$ be the subset of $\mc N_u$ corresponding to users that are of the same type as $u$ and to which additionally an recommendation has been made by drawing $T_r$ items uniformly from $\setQ$ for each user $u$. 
Let $N_g$ be the cardinality of  $\mc N_u^{\text{good}}$. 
In order to upper-bound $\PR{\hat p_{ui} \leq \frac{\pf}{2} }$, we first note that by~\eqref{eq:events}, 
\begin{align}
\PR{\hat p_{ui} \leq \frac{\pf}{2} }
\leq  
\PR{\hat p_{ui} \leq \frac{\pf}{2} \Big| N_g \geq n_g }
+ \PR{N_g \leq n_g}.
\label{eq:hatpuistart}
\end{align}
Here, we defined 
\begin{align}
n_g \defeq \frac{T_r k}{Q} (1/2- \beta). 
\label{eq:choiceng}
\end{align}
We next upper bound the probabilities on the RHS of \eqref{eq:hatpuistart}. 
We start with the first probability on the RHS of \eqref{eq:hatpuistart}: 
\begin{align}
\PR{ \hat p_{ui} 
\leq \frac{\pf}{2} \Big| N_g =  n_g' }
\hspace{-2.5cm}&\hspace{2.5cm}\leq
\PR{ \frac{\sum_{v \in \mc N_u^{\text{good}} }  \ocR_{vi} }{ n_g' + \nb} \leq \frac{\pf}{2} \Big| N_g =  n_g'} \label{eq:NbNgpro} \\
&= 
\PR{ \frac{1}{n_g'}   \sum_{v \in \mc N_u^{\text{good}} }  \ocR_{vi}
 \leq \frac{\pf}{2} \frac{n'_g + \nb}{ n_g'}    \Big|  N_g = n_g'} \nonumber \\
&= 
\PR{ \frac{1}{n_g'}   \sum_{v \in \mc N_u^{\text{good} } } \ocR_{vi} 
 \leq  \pf\left(\frac{1}{2} + \Delta \right) -  \pf \left(\Delta - \frac{\nb}{2 n_g'} \right) 
 \Big|  N_g = n_g'
 } \nonumber \\
&=\PR{    \sum_{v \in \mc N_u^{\text{good} } }  
\left(\ocR_{vi} - \pf\left(\frac{1}{2} + \Delta \right)
\right)
 \leq   -  n_g' \pf \left(\Delta - \frac{\nb}{2 n_g'} \right) 
 \Big|  N_g = n_g'
 } \nonumber \\
&\leq  
%
%
e^{ -  \frac{  n_g' \pf  (\Delta - \nb/ (2n_g') )^2 / 2 }{  (1/2 + \Delta)  +  (\Delta - \nb/(2n_g'))/3  } }
\label{eq:boundNgNb}
\end{align}
where \eqref{eq:NbNgpro} follows from the number of users $n_{ui}$ in $\mc N_u$ that received recommendation $i$ being upper bounded by $N_g + \beta k$ (recall that $\beta k$ is the maximum number of bad neighbors in $\mc N_u$), 
and by assuming adversarially that all recommendations given to bad neighbors 
did yield $\ocR_{vi} = 0$. 
Finally, \eqref{eq:boundNgNb} follows from Bernstein's inequality; 
to apply Bernstein's inequality, we used that $\EX{\ocR_{vi}} = \pf (1/2 + \Delta)$, and that the variance of $\ocR_{vi}$ is upper bounded by $\pf (1/2 + \Delta)$, for $v\in \mc N_u^{\text{good}}$.  
%
%
Next, note that by Bayes theorem,
\begin{align}
\PR{\hat p_{ui} \leq 1/2 | N_g \geq n_g }
&=
\frac{\PR{ \{ \hat p_{ui} \leq 1/2 \} \cap \{ N_g \geq n_g \}  } }{
\PR{N_g \geq n_g }
} \nonumber \\
&= 
\frac{
\sum_{n_g' \geq n_g} 
\PR{\hat p_{ui} \leq 1/2  \big| N_g \geq n_g }
\PR{ N_g = n_g' }
}{
\PR{N_g \geq n_g }
} \nonumber \\
&\leq 
e^{ -  \frac{  n_g \pf  (\Delta - \nb/ (2n_g') )^2 / 2 }{  (1/2 + \Delta)  +  (\Delta - \nb/(2n_g'))/3  } } \label{eq:usedNbNb}\\
&\leq 
e^{ -  \frac{  n_g \pf  \Delta^2 / 8 }{  1/2 + \Delta + \Delta/6  } } 
\leq 
e^{ -  \frac{  n_g \pf  \Delta^2 }{  16  } } 
\leq 
e^{ -  \frac{  T_r k  \pf  \Delta^2 }{ Q 64  } }.
\label{eq:condhatpui}
\end{align}
Here, inequality~\eqref{eq:usedNbNb} follows from inequality~\eqref{eq:boundNgNb} and using that the RHS of inequality~\eqref{eq:boundNgNb} is increasing in $n_g'$. 
For inequality~\eqref{eq:condhatpui} we used the definition of $n_g$ in \eqref{eq:choiceng}, and that 
\begin{align}
\frac{\nb}{n_g} 
=
\frac{\beta k}{ \frac{T_r k}{Q} (1/2- \beta) }
=
\frac{Q}{T_r } 
\frac{\beta }{ 1/2 - \beta } \leq \Delta.
\label{eq:ineqfracnbng}
\end{align}
Here, the inequality \eqref{eq:ineqfracnbng} holds by $\beta \leq \frac{\Delta T_r}{4Q}$, by assumption, and $\beta \leq 1/4$, due to $\Delta \leq 1/2$ and $T_r \leq Q$ (since we recommend each item at most once).

We proceed with upper bounding $\PR{N_g \leq n_g}$ in \eqref{eq:hatpuistart}. 
Recall that $N_g$ is the number of times item $i$ has been recommended to one of the $\geq (1-\beta) k$ good neighbors in $\mc N_u$. 

We will only consider the $T_r$ random items recommended to each user; this yields an upper bound on $\PR{N_g \leq n_g}$. 
Recall that those items are chosen from the $Q$ items in $\setQ$, and that, by assumption, of the $k$ neighbors at least $(1-\beta) k$ are good. 
By Bernstein's inequality, 
%
\begin{align}
\PR{N_g \leq n_g} 
&= 
\PR{N_g \leq T_r \frac{(1-\beta)k}{Q}  - 
\frac{T_r k}{2Q}
} \nonumber \\
&\leq 
e^{- \frac{T_r k ( \frac{1}{2Q} )^2/2}{
\frac{1-\beta}{Q}(1 - \frac{1-\beta}{Q} )
+ \frac{1}{3} \frac{1}{2Q}
} }
%
\leq 
e^{- \frac{T_r k ( \frac{1}{2Q} )^2/2}{
\frac{1-\beta}{Q}(1 - \frac{1-\beta}{Q} )
+ \frac{1}{3} \frac{1}{2Q}
} } 
\leq 
e^{- \frac{T_r k \frac{1}{8Q} }{
1 + 1/6
} } 
\leq e^{- \frac{T_r k}{10 Q} }. 
\label{eq:Ngleqng} 
\end{align}
Application of inequalities~\eqref{eq:condhatpui} and \eqref{eq:Ngleqng} to inequality~\eqref{eq:hatpuistart} together with a union bound yields 
\begin{align}
\PR{\hat p_{ui} \leq 1/2, \text{for one or more } i \in \setQ}
&\leq 
Q \left(  e^{ -  \frac{  T_r k  \pf  \Delta^2 }{ Q 64  } }  + e^{-   \frac{T_r k}{10 Q} }  \right) 
\leq 
2 Q e^{ -  \frac{  T_r k  \pf  \Delta^2 }{ Q 64  } },
\end{align}
where we used that $\pf \Delta^2 \leq 1$. 
By \eqref{eq:batch}, the RHS above is smaller than $\delta'$. This concludes the proof. 

\section{Proof of Proposition~\ref{prop:necessity}}

Consider a set of users with $K$ user types that are non-overlapping in their preferences, specifically, 
consider a set of users where every user $u$ belonging to the $k$-th user type has preference vector
\[
[\vp_u]_{i}
=
\begin{cases}
1, & \text{if } i \in [ k(M-1)/K, \ldots,  k M/K ] \\
0, & \text{otherwise}.
\end{cases}
\]
Consider a given user $u$. 
At time $T$, the expected number of ratings obtained by $u$ is upper bounded by $\pf^2$. 
Thus, for all $T \leq \frac{\lambda}{\pf^2}$ in at least a fraction $\lambda$ of the runs of the algorithm, the algorithm has no information on the user $u$, and the best it can do is to recommend a random item. For our choice of preference vectors, 
with probability at most $1/K$, it will recommend a likable item. 
Therefore, an upper bound on the expected regret is given by $(\lambda + 1/K) NT$.


\end{document}